# Evaluating the Efficacy of Large Language Models in Detecting Fake News: A Comparative Analysis

Sahas Koka, Anthony Vuong, Anish Kataria
sahas.koka@gmail.com, anthuny.vuong@gmail.com, anishesg@gmail.com

## Abstract

In an era increasingly influenced by artificial intelligence, the detection of fake news is crucial, especially in contexts like election seasons where misinformation can have significant societal impacts. This study evaluates the effectiveness of various LLMs in identifying and filtering fake news content. Utilizing a comparative analysis approach, we tested four large LLMs—GPT-4, Claude 3 Sonnet, Gemini Pro 1.0, and Mistral Large—and two smaller LLMs—Gemma 7B and Mistral 7B. By using fake news dataset samples from Kaggle, this research not only sheds light on the current capabilities and limitations of LLMs in fake news detection but also discusses the implications for developers and policymakers in enhancing AI-driven informational integrity.

## I. INTRODUCTION

In today's digital age, the proliferation of fake news has become a significant problem, impacting public opinion, political decisions, and social trust. Studies indicate that fake news spreads more rapidly and widely than true news, with potentially devastating consequences. For instance, a study by MIT found that false news stories are 70% more likely to be retweeted than true stories. With the upcoming election season, the ability to accurately detect and mitigate the spread of fake news is more critical than ever.

The rise of machine learning (ML) and artificial intelligence (AI) has brought new challenges and opportunities in the fight against fake news. While AI and ML technologies have been used to create convincing fake news articles, they also offer powerful tools to identify and combat misinformation. In this context, our research seeks to explore the effectiveness of various large language models in detecting fake news.

Large Language Models (LLMs) are a type of AI model designed to understand and generate human-like text based on vast amounts of data. These models are trained on diverse datasets containing billions or even trillions of parameters, enabling them to capture intricate patterns in language and provide highly accurate and contextually relevant responses. Large LLMs are exceptionally powerful in processing and understanding complex information, making them more accurate in tasks such as fake news detection. However, they require substantial computational resources. On the other hand, smaller LLMs are faster and more efficient, making them suitable for applications where speed is crucial, though they may not be as precise as their larger counterparts.

We conducted a comparative analysis of six prominent language models: GPT-4, Mistral Large, Claude 3 Sonnet, Gemini Pro 1.0, Mistral 7B, and Gemma 7B. Utilizing the Fake News Detection Datasets by Emine Bozkus on Kaggle, which consists of over 20,000 real and fake articles, we selected a random sample of 15 real and 15 fake articles for our study. By employing the OpenRouter tool, we were able to simultaneously generate responses from each model, facilitating a straightforward comparison of their performance.

Our approach addresses the need for scalable, efficient methods to detect fake news in a landscape increasingly dominated by AI-generated content. By leveraging advanced language models, we aim to contribute to the development of robust solutions capable of identifying and mitigating the impact of fake news. Our findings highlight the varying strengths and weaknesses of these models, providing valuable insights into their potential applications in the ongoing battle against misinformation.

In this paper, we present our methodology, discuss the comparative performance of the models, and offer conclusions on their effectiveness in detecting fake news. By doing so, we aim to underscore the importance of harnessing AI and ML technologies in combating the spread of false information and protecting the integrity of public discourse.

## II. Problem Definition and Algorithm

### A. Task Definition

The problem we are addressing is the evaluation of various large language models (LLMs) for their ability to detect fake news. Specifically, we aim to determine which models are more accurate and reliable in distinguishing between real and fake news articles. This is an interesting and important problem because the proliferation of fake news has significant implications for public opinion, societal trust, and decision-making processes in both governmental institutions and research bodies. Reliable detection of fake news is crucial for maintaining the integrity of information dissemination and supporting informed decision-making.

Models Tested:
- Large language models
    - GPT-4 is a language model developed by OpenAI with 1.76 trillion parameters
    - Claude 3 is a language model by Anthropic with approximately 70 billion parameters
    - Mistral Large, developed by Mistral AI, is a language model with 65 billion parameters
    - Gemini Pro 1.0 is a language model by Google with 640 billion parameters
    - Mistral 7B, developed by Mistral AI, is a language model with 7 billion parameters
    - Gemma 7B, developed by Google DeepMind, is a language model with 7 billion parameters

Inputs:
- A set of news articles consisting of both real and fake news.

Outputs:
- Performance metrics for each model in detecting fake news, such as accuracy, precision, recall, F1 score with P-values.
- A comparative analysis highlighting the strengths and weaknesses of each model in the context of fake news detection.

Steps:

1. Data Preparation:
    a. Use the Fake News Detection Datasets by EMINE BOZKUŞ from Kaggle.
    b. Randomly select 15 real articles and 15 fake articles from the dataset.
2. Model Testing:
    a. Utilize the OpenRouter tool to generate responses from each LLM for the selected articles.
    b. Collect the responses in a spreadsheet for analysis.
3. Performance Evaluation:

        a. Evaluate the performance of each model using metrics such as accuracy, precision, recall, F1 score with P-values.
        b. Analyze the results to identify which models perform best in detecting fake news.
    4. Comparative Analysis:
        a. Compare and share the performance of the models for fake news detection tasks.

This detailed analysis will help in understanding the capabilities of different LLMs in the context of fake news detection and will guide the selection of the most suitable models for practical applications.

## III. EXPERIMENTAL EVALUATION
### A. Methodology

Methods and Statistics:

To evaluate the effectiveness of large language models in detecting fake news, we utilized a combination of statistical evaluation and case-by-case processing methods. Our approach involved using existing experimental data from the Fake News Detection Datasets by Emine Bozkus on Kaggle, which comprises over 20,000 real and fake articles. We randomly selected 15 real articles and 15 fake articles from this dataset to ensure a balanced and representative sample for our analysis.

| True.csv (53.58 MB) | | | | Fake.csv (62.79 MB) | | | |
| --- | --- | --- | --- | --- | --- | --- | --- |
| title | text | subject | date | title | text | subject | date |
| As U.S. budget fight looms, Republicans flip their fiscal script | WASHINGTON (Reuters) - The head of a conservative Republican faction in the U.S. Congress, who voted... | politicsNews | December 31, 2017 | Donald Trump Sends Out Embarrassing New Year's Eve Message; This is Disturbing | Donald Trump just couldn t wish all Americans a Happy New Year and leave it at that. Instead, he had... | News | December 31, 2017 |

Above are examples of the data we used:

We employed the OpenRouter tool to generate responses from each of the six large language models (GPT-4, Claude 3 Sonnet, Mistral Large, Gemini Pro 1.0, Mistral 7B and Gemma 7B) simultaneously. This tool facilitated the collection of response data in a structured format, allowing for straightforward comparison and analysis.

In this study, we employed zero-shot prompting to evaluate varying performances between large language models. For each article, we provided the models with the text and directly asked whether the article was real or fake. This approach ensured that the models had to rely on their pre-existing knowledge and inference capabilities without any additional training or context-specific fine-tuning, allowing for a fair and straightforward comparison of their abilities to identify misinformation.

Evaluation Criteria

Our evaluation criteria focused on several key metrics to provide a comprehensive assessment of each model's performance:

1.	Accuracy: This metric measures the proportion of correctly identified articles (both real and fake) out of the total articles evaluated. Accuracy gives a straightforward indication of overall performance.

2.	Precision: Precision is the ratio of true positive identifications (correctly identified fake news) to the total number of positive identifications (both true and false positives) made by the model. High precision indicates that the model makes few false positive errors.

3.	Recall: Recall, also known as sensitivity, is the ratio of true positive identifications to the total number of actual positive cases (true positives and false negatives) in the dataset. High recall indicates that the model effectively identifies most of the fake news articles.

4.	F1 Score: The F1 score is the harmonic mean of precision and recall, providing a single metric that balances both concerns. It is particularly useful when there is an uneven class distribution, such as having more real news than fake news.

5.	P-value: A p-value is a statistical measure that helps researchers determine the significance of their results, indicating the probability of observing the tested effect or one more extreme if the null hypothesis is true. Discrepancies in p-values can occur due to issues like sample size, data variability, or errors in data handling or analysis methods, which might affect the accuracy and reliability of the statistical conclusions.

## Hypothesis

Our experiment tested the following hypotheses:

1. Hypothesis 1: Large language models can accurately detect fake news with a high degree of precision and recall.
2. Hypothesis 2: There are significant differences in performance among the six large language models tested. This hypothesis addresses the need to understand which models perform better and under what conditions.

For each article in our sample, we inputted the text into the OpenRouter tool, which generated responses from each model. We then evaluated each response based on whether the model correctly identified the article as real or fake. This process ensured a fair comparison across all models.

## Data Collection

The results were compiled into a spreadsheet for detailed analysis. We used statistical methods, including confusion matrices and performance metrics, to compare the effectiveness of each model. By presenting and analyzing this data, we aim to provide a comprehensive evaluation of the capability of large language models to detect fake news. Our findings offer insights into their practical applications and limitations, contributing to the ongoing efforts to combat the spread of misinformation.

# B. Results

*Quantitative Analysis of AI Model Performance*

The analysis of the AI models ability to detect fake news involved assessing their accuracy in identifying both real and fake news. The results showed high accuracy rates across all models.

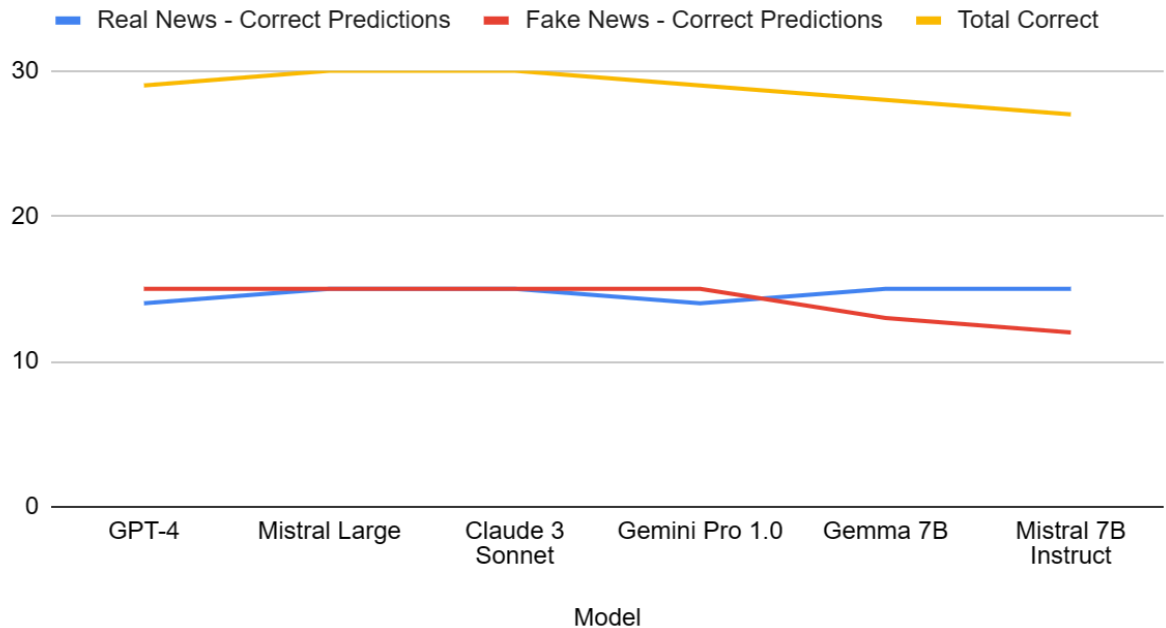

|  | Accuracy | Precision | Recall | F1 Score | P-value* |
|---|---|---|---|---|---|
| **GPT4** | 0.967 | 1 | 0.933 | 0.9654 | 0.163 |
| **Mistral Large** | 1 | 1 | 1 | 1 | N.A( Reference Distribution) |
| **Claude 3 Sonnet** | 1 | 1 | 1 | 1 | N.A( Reference Distribution) |
| **Gemini Pro 1.0** | 0.966 | 1 | 0.933 | 0.9654 | 0.163 |
| **Gemma 7B** | 0.9333 | 0.882 | 1 | 0.937 | 0.08 ( Stat sign < 0.1) |
| **Mistral 7B Instruct** | 0.9 | 0.833 | 1 | 0.909 | 0.042 ( Stat sign < 0.1) |

*Mistral Large and Claude 3 Sonnet results are used as reference distribution for P-value computation.

*Overall Performance:*

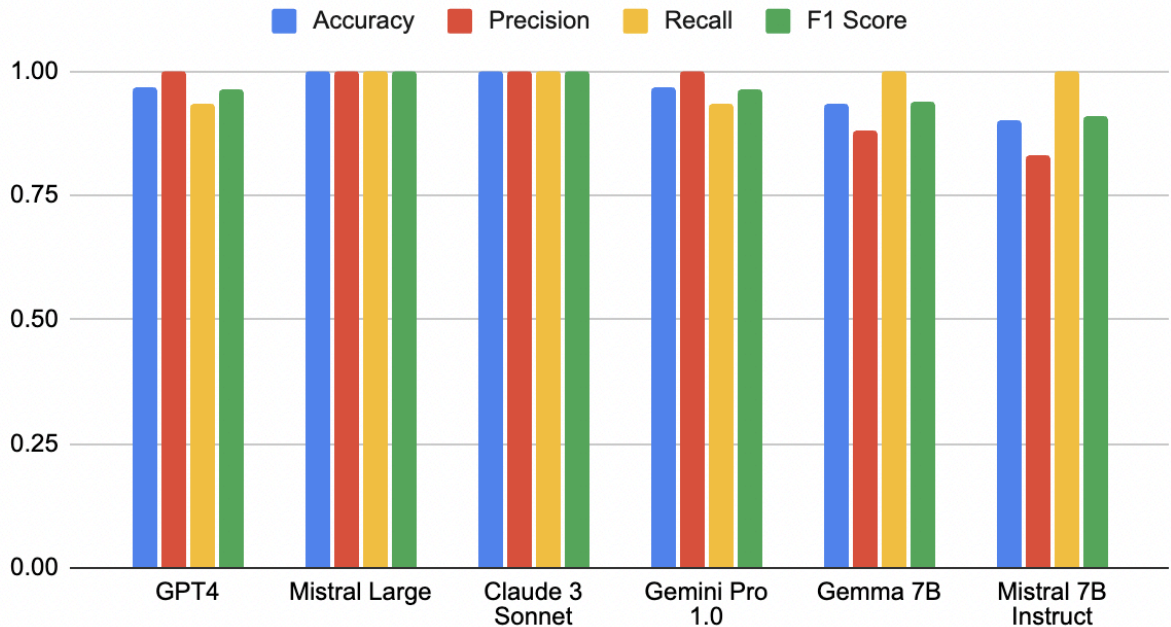

Large models like GPT-4, Mistral Large, Claude 3 Sonnet and Gemini Pro 1.0 consistently demonstrated superior performance across all metrics compared to smaller models such as Gemma 7B and Mistral 7B Instruct. This trend highlights the advantage of increased parameters and model complexity in processing and accurately identifying nuanced patterns in data.

*Accuracy:*

Claude 3 Sonnet and Mistral Large achieved the highest accuracy at 1 indicating their robust capability to correctly classify both real and fake news. Other large models GPT4 and Gemini Pro 1.0 accuracies are at 0.97 which is not statistically significant with respect to Claude 3 Sonnet and Mistral Large. Smaller models like Gemma 7B and Mistral 7B Instruct showed lower accuracy and the difference is statistically significant with respect to Claude 3 Sonnet and Mistral Large. Mistral 7B Instruct has the lowest accuracy of 0.9 among all models.

*Precision and Recall:*

All models maintained a high balance between precision and recall, with most achieving a perfect score in at least one of these metrics. All Large models had perfect precision, meaning they made very few false positive errors. Smaller models showed some variability, with Mistral 7B having a lower precision (0.833) but perfect recall (1), suggesting it was better at identifying fake news but more prone to false positives.

*F1 Score:*

The F1 Score further confirmed the superior performance of large models, combining both precision and recall into a single metric. Large models like Mistral Large and Claude 3 Sonnet achieved perfect F1 scores, whereas smaller models lagged slightly behind.

*P-value:*

The P-value analysis revealed that Large LLMs like Mistral Large and Claude 3 Sonnet results are statistically better than small LLMs like Gemma 7B and Mistral 7B at a confidence interval of 90%.

Overall, larger models like GPT-4, Claude 3 Sonnet, Gemini Pro 1.0, and Mistral Large demonstrated superior performance in fake news detection compared to smaller models such as Gemma 7B and Mistral 7B Instruct. The larger models consistently achieved higher accuracy, precision, recall, and F1 scores, indicating they are more reliable and effective in distinguishing between real and fake news. The smaller models, while still effective, showed more variability and lower precision, suggesting a higher likelihood of false positive errors. These findings support Hypothesis 1, affirming the capability of large language models to accurately detect fake news with high precision and recall, and Hypothesis 2, highlighting significant performance differences among the tested models.

## C. Discussion

Hypothesis 1: Accuracy of Large Language Models in Detecting Fake News
Our research demonstrates that large language models (LLMs) such as GPT-4, Claude 3 Sonnet, Gemini Pro 1.0, and Mistral Large exhibit a high degree of precision and recall in detecting fake news. In a comparative analysis using 15 real and 15 fake news articles, these models achieved impressive accuracy scores ranging from 0.967 to 1.0. Their precision scores were consistently perfect (1.0), indicating minimal false positive errors, and their recall scores were near-perfect or perfect, ensuring effective identification of fake news articles. The F1 scores further supported their balanced performance. In contrast, smaller models like Gemma 7B and Mistral 7B Instruct, while still effective, showed lower precision and F1 scores, suggesting a higher likelihood of false positives. These findings validate our hypothesis that large LLMs can accurately detect fake news with a high degree of precision and recall, highlighting their potential as reliable tools in combating misinformation.

Hypothesis 2: Variance in Performance Among Models
Our study reveals significant differences in performance among the six large language models tested for fake news detection. Larger models like GPT-4, Claude 3 Sonnet, Gemini Pro 1.0, and Mistral Large consistently outperformed smaller models such as Gemma 7B and Mistral 7B Instruct. The larger models achieved higher accuracy, precision, recall, and F1 scores, with Mistral Large and Claude 3 Sonnet attaining perfect scores across all metrics. In contrast, the smaller models exhibited more variability, with lower precision and F1 scores, indicating a higher likelihood of false positives. These findings underscore the superior performance of larger models in accurately detecting fake news and highlight significant differences in model capabilities.

**Real World Implications:**

1. Influence on Education
The findings of this study underscore the necessity of integrating artificial intelligence and machine learning into educational curriculums to better prepare students for a digitally dominated world. By exposing students to the capabilities and limitations of large language models (LLMs) in detecting fake news, educators can cultivate digital literacy skills. This knowledge is particularly valuable in an era where information consumption predominantly occurs through digital media. Students trained in understanding and leveraging AI for information verification can better navigate the complexities of modern information ecosystems, thus enhancing their decision-making skills and promoting a more informed citizenry.

2. Influence on Business
For businesses, particularly those in the digital content, media, and technology sectors, the effective detection of fake news using advanced LLMs can serve as a crucial competitive advantage. Companies can employ these models to ensure the integrity of the content they disseminate, thereby boosting consumer trust and satisfaction. Furthermore, businesses that develop or utilize AI tools can leverage these findings to refine their products, leading to more reliable and efficient services. This can translate to improved reputation management and operational efficiencies, as businesses are better equipped to mitigate the risks associated with the spread of misinformation.

3. Influence on Jobs
The advancement in AI capabilities for detecting fake news as demonstrated by the study suggests a shift in job roles within industries reliant on information validation, such as journalism, social media management, and content moderation. The reliance on AI to filter and verify information could lead to the evolution of these roles, where the emphasis might shift towards supervising AI operations and managing responses to AI-generated insights. Additionally, this shift could spur the creation of new career paths focused on the development, maintenance, and ethical management of AI systems. Professionals with expertise in AI and machine learning, particularly in the context of language models and misinformation detection, are likely to be in high demand as businesses and governments seek to harness these technologies to secure the informational landscape.

# IV. Related Work

In this section, we compare our research with several other significant studies in the field of fake news detection using machine learning techniques. The comparison is based on the problem each study addresses, their methodologies, and how our approach provides improvements or differences.

"Fake News Detection using Machine Learning Algorithms" by Sharma et al.
Problem and Method: This study by Sharma et al. focuses on the classification of online news articles as fake or real using machine learning algorithms such as Naive Bayes, Random Forest, and Logistic Regression. The dataset used includes articles from various online platforms, and the study emphasizes the rapid spread of fake news via social media and its impact on public opinion and political outcomes.
- Differences: Our study differs in its use of multiple large language models (LLMs) like GPT-4, Gemma 7B, and others to detect fake news. We cross-tested these models using a specific dataset from Kaggle and used OpenRouter to automate the response collection process from the models simultaneously.
- Improvements: Our method introduces a comparative analysis of multiple LLMs, which is a step forward in evaluating the capabilities of state-of-the-art AI models in detecting fake news. This approach provides insights into the strengths and weaknesses of various LLMs in handling misinformation.

"Fake News Detection Using Machine Learning Approaches" by Z Khanam et al.

Problem and Method: Khanam et al. propose a supervised machine learning model to classify news articles as true or false based on textual analysis. They use various classifiers like Naive Bayes, SVM, and Random Forest, focusing on the precision of these models in detecting fake news.
- Differences: Similar to Sharma et al., Khanam et al. employ traditional machine learning classifiers. Our study extends this by integrating advanced LLMs, offering a broader perspective on the capabilities of contemporary AI technologies.
- Improvements: By incorporating advanced LLMs and comparing their performance, our research provides a more nuanced understanding of how modern AI tools can enhance fake news detection, potentially outperforming traditional classifiers in terms of accuracy and contextual understanding.

By comparing these works, our research highlights the progression from traditional machine learning techniques to the application of advanced LLMs in the fight against fake news. The empirical results from our study provide a valuable contribution to the field, demonstrating the potential of LLMs in enhancing the accuracy and reliability of fake news detection systems.

# V. Future Work

In planning future research, it is crucial to address several shortcomings identified in the current methodology to enhance the effectiveness and applicability of fake news detection systems. One major limitation is the relatively small dataset of 30 articles used in the study. To develop a more robust and universally applicable model, future research should include a larger and more diverse dataset. Expanding the dataset will help in capturing a wider variety of linguistic features and news formats, thereby improving the generalizability of the findings.

Additionally, while the current study provided valuable insights, it was restricted to a specific set of language models. As the field of artificial intelligence is rapidly evolving, continuously updating the set of models tested will be vital. Future studies should explore the use of newer and potentially more powerful models as they become available. Moreover, incorporating ensemble methods that integrate outputs from multiple models could potentially boost both accuracy and reliability in detecting fake news.

Another area for improvement involves the analysis process itself. The reliance on automated tools like OpenRouter, while beneficial for consistency and scalability, may not fully exploit each model's capabilities. Future methodologies should consider manual fine-tuning of models based on specific detection tasks, which could optimize performance tailored to the nuanced characteristics of fake and real news. Furthermore, moving beyond binary classification and adopting a multi-class approach to categorize various types of misinformation could provide deeper insights and more effective detection strategies. This nuanced approach would allow for more detailed analysis and better handling of complex misinformation types such as satire or misleading content that may not be outright fake but still harmful.

Lastly, integrating contextual and source credibility analysis can significantly improve the detection process. Many sophisticated fake news articles are crafted with convincing contexts or mimic reputable sources. Analyzing the source credibility and the context in which information is presented will enhance the ability to discern true news from fake news effectively. These guidelines not only aim to address the current study's limitations but also pave the way for more comprehensive and accurate fake news detection systems in future research endeavors.

# VI. Conclusion

This study has provided a comprehensive evaluation of the efficacy of various large language models (LLMs) in detecting fake news. Among the models tested, those with a higher number of parameters, such as GPT-4, Mistral Large, Claude 3 Sonnet, and Gemini Pro 1.0, generally demonstrated superior accuracy, precision, and recall compared to their smaller counterparts. This finding supports the hypothesis that LLMs with more parameters can more effectively process and analyze complex textual data, thereby enhancing their capability to identify fake news.

Our comparative analysis shows that while there are minor performance variations among the models, the models with more parameters tend to outperform those with fewer in terms of nuanced text understanding and misinformation detection. This underscores the potential of high-parameter LLMs to lead advancements in the accuracy and reliability of fake news detection systems.

However, the study also highlights the need for larger and more diverse datasets to fully challenge and exploit the capabilities of these advanced models. Future research should focus on expanding the dataset size and complexity, which will not only test these models more rigorously but also help in refining their architecture and training processes.

In conclusion, the results from this study indicate that LLMs with more parameters are highly effective tools for combating misinformation. Continued advancements in model design and training, coupled with more challenging test conditions, will pave the way for developing more robust, efficient, and accurate fake news detection systems.

Authors:

**First Author** - Sahas Koka, sahas.koka@gmail.com
**Second Author -** Anthony Vuong, anthuny.vuong@gmail.com
**Third Author -** Anish Kataria, anishesg@gmail.com